\documentclass{article}

\usepackage{times}
\usepackage{graphicx} % more modern
\usepackage{subfigure} 

\PassOptionsToPackage{numbers, compress}{natbib}

\usepackage{algorithm}
\usepackage{algorithmic}

\usepackage{helvet}
\usepackage{courier}

\usepackage{makecell}
\usepackage{graphicx}
\usepackage{subfigure}
\usepackage{color}
\usepackage{amsfonts}
\usepackage{amsmath}
\usepackage{amssymb}

\def\etal{{\em et al.\/}\, }

\def\ie{\textit{i.e.}}
\def\eg{\textit{e.g.}}

\DeclareMathAlphabet\mathbfcal{OMS}{cmsy}{b}{n}
\def\0{{\bf 0}}
\def\1{{\bf 1}}

\def\bA{{\bf A}}

\def\bW{{\bf W}}
\def\bX{{\bf X}}

\def\bZ{{\bf{Z}}}

\def\bX{{\bf X}}

\def\bW{{\bf W}}

\usepackage[final]{neurips_2019}
\usepackage{booktabs}
\usepackage[utf8]{inputenc} % allow utf-8 input
\usepackage[T1]{fontenc}    % use 8-bit T1 fonts
\usepackage{hyperref}       % hyperlinks
\usepackage{url}            % simple URL typesetting
\usepackage{booktabs}       % professional-quality tables
\usepackage{amsfonts}       % blackboard math symbols
\usepackage{nicefrac}       % compact symbols for 1/2, etc.
\usepackage{microtype}      % microtypography
\usepackage{multirow}
\usepackage{xspace}
\usepackage{enumitem}

\usepackage{wrapfig}

\def\qicam{\textcolor{black}}
\def\guo{\textcolor{black}}
\def\guocam{\textcolor{black}}

\definecolor{amethyst}{rgb}{0.6, 0.4, 0.8}

\def\kuired{\textcolor{black}}
\definecolor{chenyaofocolor}{RGB}{39, 174, 96}

\definecolor{Purple}{RGB}{128, 0, 128}

\definecolor{xuskcolor}{RGB}{255, 153, 51}

\def\mytitle{NAT: Neural Architecture Transformer for Accurate and Compact Architectures}
\newcommand{\eat}[1]{}

\newcommand{\sexyname}{NAT\xspace}

\title{\mytitle}

\author{
	Yong Guo\thanks{Authors contributed equally.}~, Yin Zheng$^*$, Mingkui Tan$^{*\dagger}$, Qi Chen, \\
	\textbf{Jian Chen\thanks{Corresponding author.}~, Peilin Zhao, Junzhou Huang}\\
	South China University of Technology, Weixin Group, Tencent, \\Tencent AI Lab, 
	University of Texas at Arlington\\
	\{guo.yong, sechenqi\}@mail.scut.edu.cn, \{mingkuitan, ellachen\}@scut.edu.cn, \\
	\{yinzheng, masonzhao\}@tencent.com, jzhuang@uta.edu
}

\begin{document}

\maketitle

\footnotetext{This work is done when Yong Guo works as an intern in Tencent AI Lab.}

\begin{abstract}
\kuired{Designing effective architectures is one of the key factors behind the success of deep neural networks.} Existing deep architectures are either manually designed or automatically searched by some Neural Architecture Search (NAS) methods.
 However, even a well-searched architecture may still contain many non-significant or redundant modules or operations (\eg, convolution or pooling),
 which \kuired{may} not only incur substantial memory consumption and computation cost \kuired{but  also deteriorate} the performance. 
 Thus, it is necessary to optimize the operations inside an architecture to improve the performance without introducing extra computation cost.
 \kuired{Unfortunately,} such a constrained optimization problem is NP-hard.  \kuired{To make the problem feasible,} we cast the optimization problem into a Markov decision process (MDP) and \kuired{seek to} learn a Neural Architecture Transformer (NAT) to replace the redundant operations with the more computationally efficient ones (\eg, skip connection or directly removing the connection). \kuired{Based on MDP, we learn NAT by exploiting reinforcement learning} to obtain the optimization policies w.r.t. different architectures.  \kuired{To verify the effectiveness of the proposed strategies, we apply NAT on both hand-crafted architectures and NAS based architectures}. \kuired{Extensive experiments on two benchmark datasets, \ie, CIFAR-10 and ImageNet, demonstrate that the transformed architecture by NAT significantly outperforms both its original form and those architectures optimized by existing  methods.}
\end{abstract}

\section{Introduction}

Deep neural networks (DNNs)~\citep{lecun1989backpropagation} have been producing state-of-the-art results in many challenging tasks including image classification~\cite{guo2018double,krizhevsky2012imagenet,srivastava2015training,zheng2014topic,zheng2015deep,Jiang:2017:VDE:3172077.3172161,guo2016shallow,zhang2019whole}, face recognition~\citep{schroff2015facenet,sun2015deeply,zheng2015neural}, 
% object detection~\citep{ren2015faster}, 
brain signal processing~\cite{nam2018brain,pan2016eeg},
video analysis~\citep{zeng2019graph,zeng2019breaking} and many other areas~\cite{zheng2016neural,zheng2016implicit,lauly2017document,cao2018adversarial,guo2019auto,guo2018dual,cao2019multi}. One of the key factors behind the success lies in the innovation of neural architectures, such as VGG~\cite{simonyan2014very} and
% GoogLeNet~\cite{szegedy2015going}, 
ResNet\cite{he2016deep}.
However, designing \kuired{effective neural architectures is often labor-intensive and relies heavily on substantial human expertise.} 
\kuired{Moreover, the human-designed process cannot fully explore the whole architecture space and thus the designed architectures may not be optimal.}
Hence, there is a growing interest to \kuired{replace the manual process of architecture design with Neural Architecture Search (NAS).}

Recently, substantial studies~\cite{liu2018darts,pham2018efficient,zoph2016neural} have shown that the automatically discovered architectures are able to achieve highly competitive performance compared to existing hand-crafted architectures. However, there are some limitations in NAS based architecture design methods. In fact, since there is an extremely large search space~\cite{pham2018efficient,zoph2016neural} (\eg, billions of candidate architectures), these methods often produce sub-optimal architectures, leading to limited representation performance or \kuired{substantial computation cost.}
\kuired{Thus, even for a well-designed model, it is necessary yet important to optimize its architecture  (\eg, removing the redundant operations) to achieve better performance and/or reduce the computation cost.}

\begin{figure*}[t]
	\centering
	\includegraphics[width=1\columnwidth]{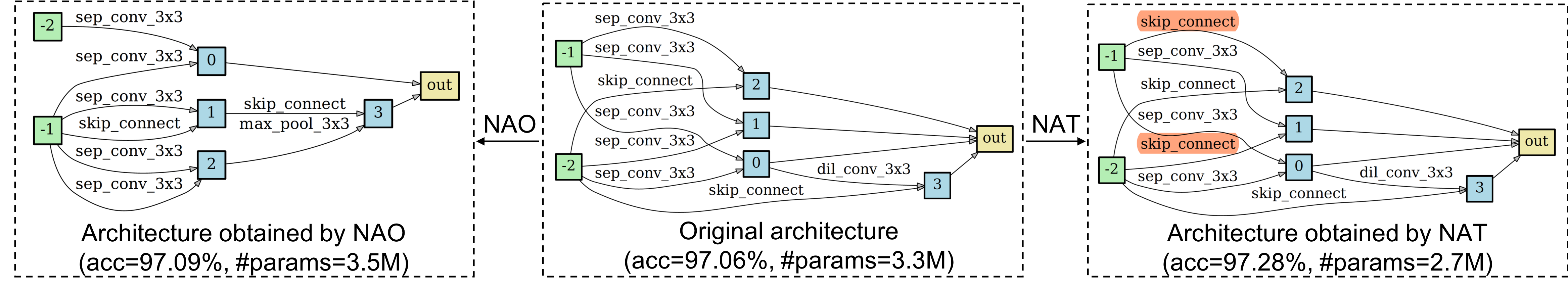}
	\caption{Comparison between Neural Architecture Optimization (NAO)~\cite{luo2018neural} and our Neural Architecture Transformer (\sexyname). Green blocks denote the two input nodes of the cell and blue blocks denote the intermediate nodes. Red blocks denote the connections that are changed by \sexyname. The accuracy and the number of parameters are evaluated on CIFAR-10 models. 
	}
% 	\vspace{-10 pt}
	\label{fig:example}
\end{figure*}

To optimize the architectures, Luo \emph{et al.} recently  proposed a neural architecture optimization (NAO) method~\cite{luo2018neural}. Specifically, NAO first encodes an architecture into an embedding in continuous space and then conducts gradient descent to obtain a better embedding. After that, it uses a decoder to \kuired{map the embedding back to obtain an optimized architecture}. However, NAO comes with its own set of limitations. First, NAO often produces a totally different architecture from the input one and may introduce extra parameters or additional \kuired{computation} cost. Second, similar to the NAS based methods,  \kuired{NAO has a huge search space, which, however, may not be necessary for the task of architecture optimization and may make the optimization problem very expensive to solve. An illustrative comparison \guocam{between} our method and NAO can be found in Figure~\ref{fig:example}. }

Unlike existing methods that design neural architectures, we seek to design an architecture optimization method, called  Neural Architecture Transformer (\sexyname), to optimize neural architectures.
% \kuired{In this paper, we propose a novel method, called  Neural Architecture Transformer (\sexyname), to
% optimize neural architectures.}
Since the optimization problem is non-trivial to solve,
we cast \kuired{it} into a Markov decision process (MDP). \kuired{Thus, the architecture optimization process} is reduced to a series of decision making problems. \kuired{Based on MDP}, we seek to replace the \kuired{expensive operations or redundant modules in the architecture with more computationally efficient ones. 
Specifically, \sexyname \kuired{shall} remove the redundant modules or replace these modules with skip connections.} 
In this way, the search space can be significantly reduced. 
\kuired{Thus, the training complexity to learn an architecture optimizer is smaller than those NAS based methods, \eg, NAO.} Last, it is worth mentioning that our \sexyname model can be used as a general architecture optimizer which takes any architecture as the input and output an optimized one.
 In experiments, we apply \sexyname to both hand-crafted and NAS based architectures and \kuired{demonstrate} the performance on two benchmark datasets, namely CIFAR-10~\citep{krizhevsky2009learning} and ImageNet~\citep{deng2009imagenet}.

% To show the effectiveness of the proposed method,
% \kuired{Given any architecture, the optimized architecture by \sexyname significantly outperform} its original form without introducing extra computation cost.

The main contributions of this paper are summarized as follows.

\begin{itemize}[leftmargin=*]

\item We propose a novel architecture optimization method, called Neural Architecture Transformer (\sexyname), to optimize arbitrary architectures in order to achieve better performance and/or \qicam{reduce} computation cost. 
To this end, \sexyname either removes the redundant paths or replaces the original operation with skip connection to improve the architecture design.

\item We cast the architecture optimization problem into a Markov decision process (MDP), 
in which we seek to solve a series of decision making problems to optimize the operations. We then solve the MDP problem with policy gradient. 
To better exploit the adjacency information of operations in an architecture, we propose to exploit graph convolution network (GCN) to build the architecture optimization model.

\item Extensive experiments demonstrate the effectiveness of our \sexyname on both hand-crafted and NAS based architectures. Specifically, for hand-crafted models (\textit{\eg,} VGG), our \sexyname automatically introduces additional skip connections into the plain network and results in 2.75\% improvement in terms of Top-1 accuracy  on ImageNet. For NAS based models (\textit{\eg,} DARTS~\cite{liu2018darts}), \sexyname reduces 20\% parameters and achieves 0.6\% improvement in terms of Top-1 accuracy on ImageNet. 

\end{itemize}

\section{Related Work}

\textbf{Hand-crafted architecture design.}
Many studies focus on architecture design and propose a series of deep neural architectures, such as Network-in-network~\cite{lin2013network}, VGG~\cite{simonyan2014very}
%GoogLeNet~\cite{szegedy2015going} 
and so on. 
Unlike these plain networks that only contain a stack of convolutions, He \emph{et al.} propose the residual network (ResNet)~\cite{he2016deep} by introducing residual shortcuts between different layers. 
However, the human-designed process often requires substantial human effort and cannot fully explore the whole architecture space, making the hand-crafted architectures often not optimal.

\textbf{Neural architecture search.}
\kuired{Recently,} neural architecture search (NAS) methods have been proposed
to automate the process of architecture design~\cite{zoph2016neural, zoph2018learning, pham2018efficient, baker2016designing, zhong2018practical,liu2018darts,cai2018proxylessnas,vaswani2017attention,so2019evolved}.
Some researchers conduct architecture search by modeling the architecture as a graph~\cite{zhang2018graph,jin2018learning}.
Unlike these methods, DSO-NAS~\cite{zhang2018single} finds the optimal architectures by starting from a fully connected block and then imposing sparse regularization~\cite{huang2018data,tan2014towards} to prune useless connections.
Besides, Jin \emph{et al.} propose a Bayesian optimization approach~\cite{jin2018efficient} to morph the deep architectures by inserting additional layers, adding more filters or introducing additional skip connections. 
More recently, Luo \emph{et al.} propose the neural architecture optimization (NAO)~\cite{luo2018neural} method to perform the architecture search on continuous space by exploiting the encoding-decoding technique. 
However, NAO is essentially designed for architecture search and often obtains very different architectures from the input architectures and may introduce extra parameters. Unlike these methods, our method is able to optimize architectures without introducing extra computation cost (See the detailed comparison in Figure~\ref{fig:example}).
    
\textbf{Architecture adaptation and model compression.}
Several methods~\cite{yang2018netadapt,dai2019chamnet,chen2015net2net} have been proposed to obtain compact architectures
by learning the optimal settings of each convolution, including kernel size, stride and the number of filters.
To obtain compact models, model compression methods~\cite{li2016pruning,he2017channel,luo2017thinet,zhuang2018discrimination} detect and remove the redundant channels from the original models. However, these methods only change the settings of convolution but ignore the fact that adjusting the connections in the architecture could be more critical. 
\guo{
Recently, Cao \etal propose an automatic architecture compression method~\cite{cao2019learnable}. However, this method has to learn a compressed model for each given pre-trained model and thus has limited generalization ability to different architectures.
Unlike these methods, we seek to learn a general optimizer for any arbitrary architecture.
% to optimize the connections in the architecture rather than the setting of each convolution.
}

\section{Neural Architecture Transformer}\label{sec:NAD}

\begin{figure*}[t]
	\centering
	\subfigure[]{
		\includegraphics[width=0.26\columnwidth]{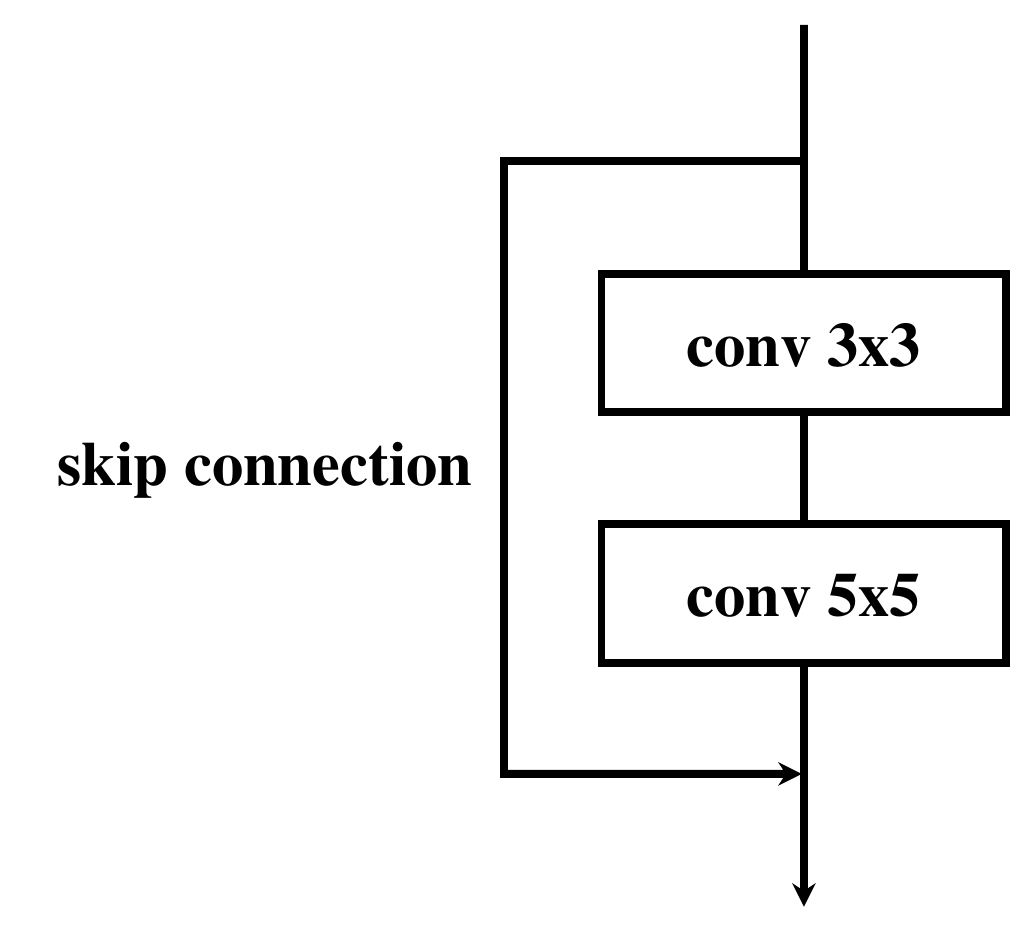}\label{fig:eg_orig_arch}
	}~~~~~~
	\subfigure[]{
		\includegraphics[width=0.26\columnwidth]{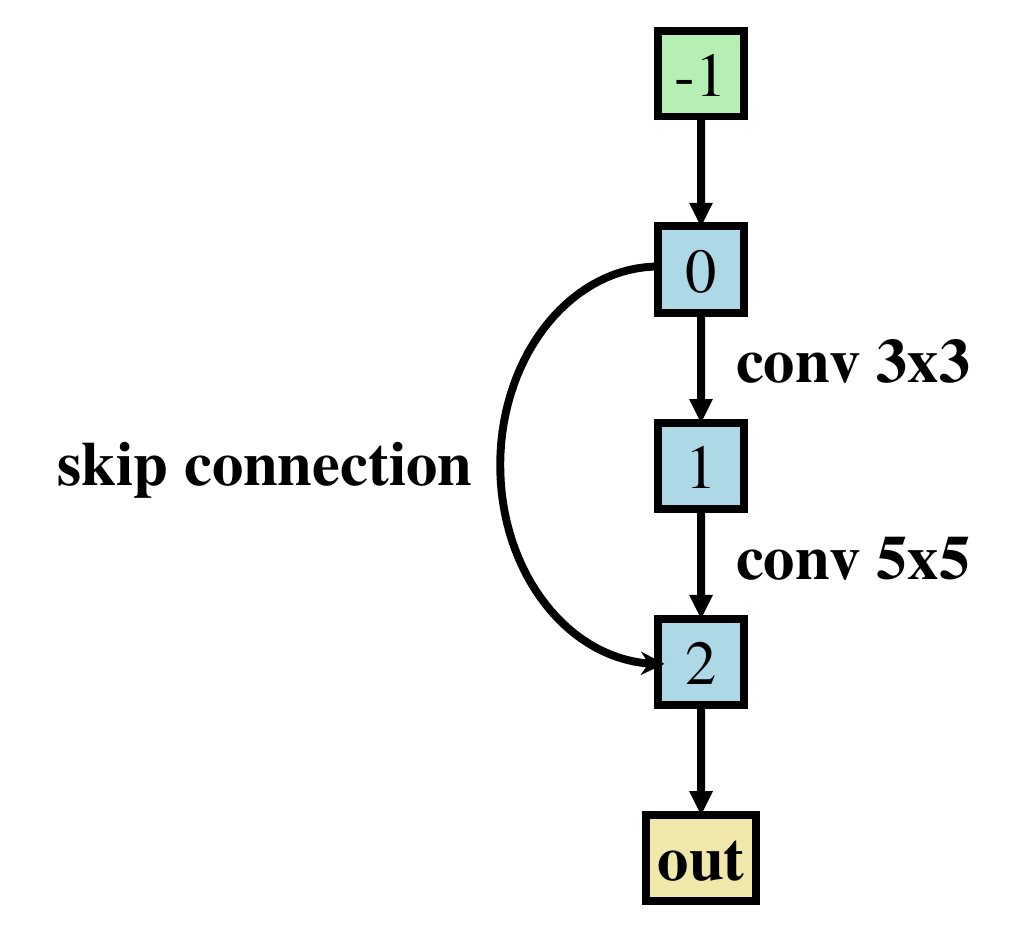}\label{fig:eg_pruned_arch}
	}~~~~~~
	\subfigure[]{
		\includegraphics[width=0.24\columnwidth]{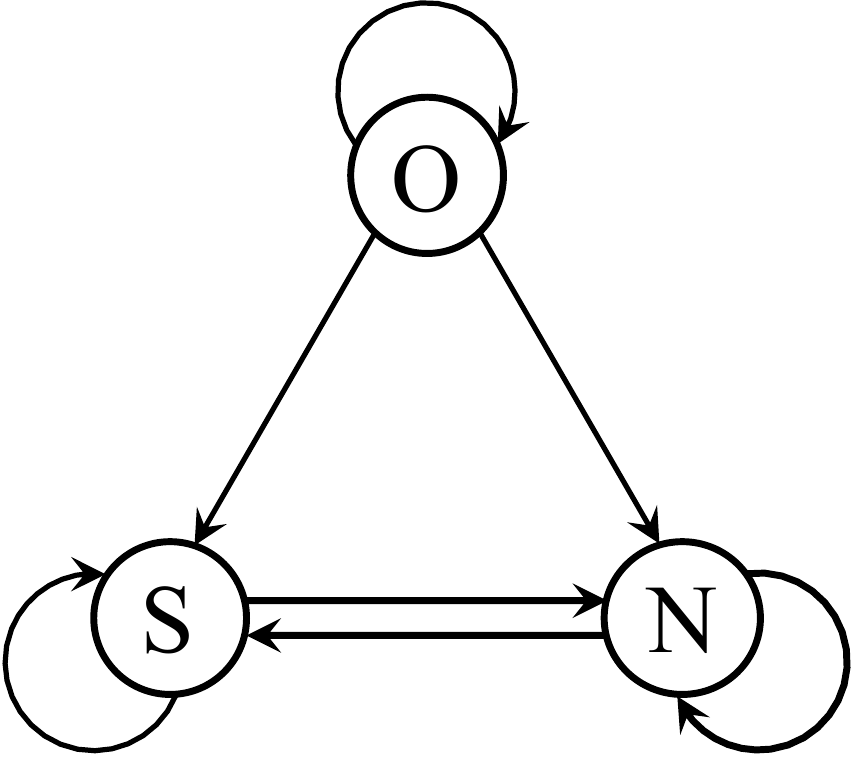}\label{fig:state_transition}
	}
% 	\vspace{-5pt}
	\caption{An example of \guocam{the} \kuired{graph representation of a residual block and the diagram of operation transformations.} (a) a residual block~\cite{he2016deep}; (b) a graph view of residual block; (c) \kuired{transformations among three kinds of operations.} ${N}$ denotes a null operation without any computation, ${S}$ denotes a skip connection, and ${O}$ denotes some computational modules other than null and skip \kuired{connections.}}
	\label{fig:arch2graph}
\end{figure*}

\subsection{Problem Definition}

Given an architecture space $\Omega$, we can represent an architecture $\alpha$ as a directed acyclic graph (DAG), \ie, $\alpha=\left(\mathcal{V},\mathcal{E}\right)$, where ${\mathcal{V}}$ is a set of nodes that denote the feature maps in DNNs and $\mathcal{E}$ is an edge set~\cite{zoph2016neural,pham2018efficient,liu2018darts}, as shown in Figure~\ref{fig:arch2graph}. 
Here, the directed edge $e_{ij} \in {\mathcal{E}}$ denotes some operation (\textit{\eg,} convolution or max pooling) that transforms the feature map from node $v_i$ to $v_j$. 
For convenience, we divide the edges in ${\mathcal{E}}$ into three categories, namely, $S$, ${N}$, $O$, as shown in Figure~\ref{fig:state_transition}.  Here, ${S}$ denotes the skip connection, ${N}$ denotes the {null connection} (\ie, no edge between two nodes), and ${O}$ denotes the operations other than skip connection or null connection (\eg, convolution or max pooling). Note that different operations have different costs. Specifically, let $c(\cdot)$ be a function to evaluate the computation cost. Obviously, we have $c({O}) > c({S}) > c({N})$.   

In this paper, we seek to design an architecture optimization method, called Neural Architecture Transformer (NAT), to optimize any given architecture into a better one with the improved performance and/or less computation cost. To achieve this, an intuitive way is to make the original operation with less computation cost, \eg, using the skip connection to replace convolution or using the null connection to replace skip connection.  Although the skip connection has slightly higher cost than the null connection, it often can significantly improve the performance~\cite{he2016deep,he2016identity}. 
Thus, we enable the transition from null connection to skip connection to increase the representation ability of deep networks. In summary, we constrain the possible transitions among $O$, $S$, and ${N}$ in Figure~\ref{fig:state_transition} in order to reduce the computation cost.

\kuired{Note that the architecture optimization} on an entire network is still very computationally expensive. \kuired{Moreover, we hope to learn a general architecture optimizer.}
Given these two concerns, we consider learning a computational cell as the building block of the final architecture. To build a cell, we follow the same settings as that in ENAS~\cite{pham2018efficient}. 
% Specifically, each cell has two input nodes that come from the outputs of previous cells, \ie, $v_{-2}$ and $v_{-1}$ and the other nodes take two previous nodes in this cell as inputs (See examples in Figure~\ref{fig:example}).
Specifically, each cell has two input nodes, \ie, $v_{-2}$ and $v_{-1}$, which denote the outputs of the second nearest and the nearest cell in front of the current one, respectively. 
Each intermediate node (marked as the blue box in Figure~\ref{fig:example}) also takes two previous nodes in this cell as inputs.  Last, based on the \kuired{learned cell}, we are able to form any final network.

\subsection{Markov Decision Process for Architecture Optimization} \label{sec:MDP}

% We show some examples in Figure~\ref{fig:example}.}

% Let $\alpha$ and $\beta$ be the variables that represent any architecture.

%Let $\beta$ be any input architecture and $\alpha$ be the optimized architecture.

% In this paper, we seek to learn a general architecture optimizer $\alpha = {\rm NAT} (\beta; \theta)$ parameterized by $\theta$, which transforms any architecture $\beta$ into an optimized architecture $\alpha$. Here, we assume $\beta$ follows some distribution $p(\cdot)$, \eg, multivariate uniformly discrete distribution. Note that  $p(\cdot)$ can be any other distributions.
% Let $w_{\alpha}$ and $w_{\beta}$ be the parameters of $\alpha$ and $\beta$, respectively.

% (\eg, the accuracy on validation data).
In this paper, we seek to learn a general architecture optimizer $\alpha = {\rm NAT} (\beta; \theta)$, which transforms any $\beta$ into an optimized  $\alpha$ and is parameterized by $\theta$. Here, we assume $\beta$ follows some distribution $p(\cdot)$, \eg, multivariate uniformly discrete distribution. \kuired{Let $w_{\alpha}$ and $w_{\beta}$ be the well-learned model parameters of architectures $\alpha$ and $\beta$, respectively.} We measure the performance of $\alpha$ and $\beta$ by some metric $R(\alpha, w_{\alpha})$ and $R(\beta, w_{\beta})$, \eg, the accuracy on validation data. For convenience, we define the performance improvement between $\alpha$ and $\beta$ by $R(\alpha | \beta) = R(\alpha, w_{\alpha}) - R(\beta, w_{\beta})$. 

To learn a good transformer $\alpha = {\rm NAT} (\beta; \theta)$ to optimize arbitrary $\beta$, we can maximize the expectation of performance improvement $R(\alpha | \beta)$ over the distribution of $\beta$ under a constraint of computation cost $c(\alpha) \leq \kappa$, where $c(\alpha)$ measures the cost of $\alpha$ and $\kappa$ is an upper bound of the cost. Then, the optimization problem can be written as
\begin{equation}\label{eq:objective}
    \begin{aligned}
         \max_{\theta} ~\mathbb{E}_{\beta \sim p(\cdot)} \left[ R \left( \alpha | \beta \right) \right], ~\text{s.t. } ~c(\alpha) \leq \kappa.
     \end{aligned}
     \vspace{-5 pt}
\end{equation}

\kuired{Unfortunately, it is non-trivial to directly obtain the optimal $\alpha$  given different $\beta$}. Nevertheless, following~\cite{zoph2016neural,pham2018efficient}, given any architecture $\beta$, we instead sample $\alpha$ from some well learned policy, denoted by $\pi(\cdot | \beta; \theta)$, namely $\alpha \sim \pi(\cdot|\beta; \theta)$. 
In other words, \sexyname first learns the policy and then conducts sampling from it to obtain the optimized architecture. In this sense, the parameters to be learned only exist in $\pi(\cdot|\beta; \theta)$.
% Since sampling from the learned policy is a specific form of the mapping ${\rm NAT(\beta; \theta)}$, we can also use $\theta$ to represent the parameters of $\pi(\cdot | \beta; \theta)$.
% and the mapping of ${\rm NAT}(\beta;\theta)$ can also be other forms, \eg, random search (See comparison in Section~\ref{exp:derive_inference}).
% \sexyname and $\pi$ share the same parameters $\theta$ since NAT obtains the optimized architecture by sampling from $\pi$.
To learn the policy, we solve the following optimization problem:
% However, \kuired{the optimization problem} in Eqn.~(\ref{eq:objective}) is only concentrated on one specific architecture $\widehat{\beta}$. In fact, we hope to find an optimizer to optimize any $\beta$, which can be sampled from the architecture distribution  
% $p(\cdot)$. 
% Let $\pi(\cdot|\beta; \theta)$ be the optimization policy parameterized by $\theta$ for the given architecture $\beta$.
% The optimization problem becomes
\begin{equation}\label{eq:objective-mean}
    \begin{aligned}
         \max_{\theta} ~\mathbb{E}_{\beta \sim p(\cdot)} \left[  \mathbb{E}_{\alpha \sim \pi(\cdot|\beta; \theta)} ~R \left( \alpha | \beta \right) \right], ~\text{s.t. } ~c(\alpha) \leq \kappa, ~\alpha \sim \pi(\cdot|\beta; \theta),
     \end{aligned}
\end{equation}
% where \guo{$\beta$ is a variable that represents the input architecture sampled from the distribution $p(\cdot)$.} 
\kuired{where $\mathbb{E}_{\beta \sim p(\cdot)} \left[  \cdot \right]$ and $\mathbb{E}_{\alpha \sim \pi(\cdot|\beta; \theta)} \left[  \cdot \right]$ denote the expectation operation over  $\beta$ and  $\alpha$, respectively.}  

This problem, however, is still very challenging to solve. \textbf{First}, the computation cost of deep networks can be evaluated by many metrics, such as the number of multiply-adds (MAdds), latency, and energy consumption, making it hard to find a comprehensive measure to accurately evaluate the cost.
\textbf{Second}, the upper bound of computation cost $\kappa$ in Eqn.~(\ref{eq:objective}) may vary for different cases and thereby is hard to determine.  Even if there already exists a specific upper bound, dealing with the constrained optimization problem is still a typical NP-hard problem. 
\textbf{Third}, how to compute $\mathbb{E}_{\beta \sim p(\cdot)} \left[  \mathbb{E}_{\alpha \sim \pi(\cdot|\beta; \theta)} ~R \left( \alpha | \beta \right) \right]$ remains a question.

% Here, $R \left( \alpha | \widehat{\beta} \right)$ can be the validation accuracy of the optimized architecture $\alpha$ obtained from the given architecture $\beta$ and its associated parameters $w$. 
% Moreover, the search space of $\alpha$ is related to specific cell structures which will be detailed later. 

To address the above challenges, we cast the optimization problem into an architecture transformation problem and reformulate it as a Markov decision process (MDP).
Specifically, we optimize architectures by making a series of decisions to alternate the types of different operations. Following the transition graph in Figure~\ref{fig:state_transition}, as  $c({O}) > c({S}) > c({N})$, we can naturally obtain more compact architectures than the given ones.
In this sense, we can achieve the goal to optimize arbitrary architecture without introducing extra cost into the architecture. Thus, for the first two challenges, we do not have to evaluate the cost $c(\alpha)$ or determine the upper bound $\kappa$.
For the third challenge, we estimate the expectation value by sampling architectures from $p(\cdot)$ and $\pi(\cdot|\beta; \theta)$ (See details in Section~\ref{sec:train}).
% Thus, the challenges regarding evaluating the cost of architectures can be avoided. We will provide the detailed MDP formulation as below.

% To address these challenges,  recall that we essentially focus on optimizing the internal operations, \ie, making a series of decisions to alternate and optimize different types of edges. In this sense, we reformulate the optimization problem into a Markov decision process. In this way, we do not need to directly compute the cost of the architecture. Instead, we can make the transition of edges follow the transition graph in  Figure~\ref{fig:state_transition} to reduce the cost.

\textbf{MDP formulation details.} A typical MDP~\cite{schulman2015trust} is defined by a tuple $(\mathcal{S}, \mathcal{A}, P, {R}, q, \gamma)$, 
 where $\mathcal{S}$ is a finite set of states, $\mathcal{A}$ is a finite set of actions, $P: \mathcal{S} \times \mathcal{A} \times \mathcal{S} \rightarrow \mathbb{R}$ is the state transition distribution, ${R}: \mathcal{S} \times \mathcal{A} \rightarrow \mathbb{R}$ is the reward function, $q: \mathcal{S} \rightarrow [0, 1]$ is the distribution of initial state, and $\gamma \in [0, 1]$ is a discount factor. 
 Here, we define an architecture as a state, a transformation mapping $\beta \to \alpha$ as an action. 
Here, we use the accuracy improvement on the validation set as the reward. Since the problem is \kuired{a} one-step MDP, \kuired{we can omit the discount factor $\gamma$.}
Based on the problem definition, \kuired{we transform any $\beta$ into an optimized architecture $\alpha$} with the policy $\pi(\cdot|\beta;\theta)$. Then, the main challenge becomes how to learn an optimal policy $\pi(\cdot|\beta;\theta)$. \kuired{Here, we exploit reinforcement learning~\cite{williams1992simple} to solve the problem and propose an efficient policy learning algorithm.}

%to transform the operations in an arbitrary architecture.

\textbf{Search space of NAT over a cell structure.} For a cell structure with $B$ nodes and 3 states for each edge, there are $2(B {\small -} 3)$ edges and the size of the search space w.r.t. a specific $\beta$ is $| \Omega_{\beta} | = 3^{2(B-3)}$. 
% Thus, \sexyname has a much smaller search space than NAS methods~\cite{pham2018efficient,zoph2016neural}, 
However, NAS methods~\cite{pham2018efficient,zoph2016neural} have a large search space with the size of $k^{2(B-3) } ((B-2)!)^2$, where $k$ is the number of candidate operations (\emph{\eg}, $k{=}5$ in ENAS~\cite{pham2018efficient} and $k{=}8$ in DARTS~\cite{liu2018darts}).

\subsection{Policy Learning by Graph Convolutional Neural Networks}\label{sec:graph_representation}

To learn the optimal policy $\pi(\cdot|\beta;\theta)$ w.r.t. an arbitrary architecture $\beta$,
we propose an effective learning method to optimize the operations inside the architecture. Specifically, we take an arbitrary architecture graph $\beta$ as the input and output the optimization policy w.r.t $\beta$. Such a policy is used to optimize the operations of the given architecture.
Since the choice of operation on an edge depends on the adjacent nodes and edges,
we have to consider the attributes of both the current edge and its neighbors. For this reason, we employ a graph convolution networks (GCN)~\cite{kipf2016semi} to exploit the adjacency information of the operations in the architecture.
Here, an architecture graph can be represented by a data pair $(\bA, \bX)$,
where $\bA$ denotes the adjacency matrix of the graph and $\bX$ denotes the attributes of the nodes together with their two input edges\footnote{Due to the page limit, we put the detailed representation methods in the supplementary.}.
We consider a two-layer GCN and formulate the model as:
\begin{equation}
    \bZ = f(\bX, \bA) = {\rm Softmax}\left(\bA {\rm \sigma}\left(\bA \bX \bW^{(0)}\right)\bW^{(1)}\bW^{\rm FC}\right),
\end{equation}
where $\bW^{(0)}$ and $\bW^{(1)}$ denote the weights of two graph convolution layers, $\bW^{\rm FC}$ denotes the weight of the fully-connected layer, $\sigma$ is a non-linear activation function (\eg, the  Rectified  Linear Unit  (ReLU)~\cite{nair2010rectified}), and $\bZ$ refers to the probability distribution of different candidate operations on the edges, \ie, the learned policy $\pi(\cdot|\beta; \theta)$. For convenience, we denote $\theta = \{ \bW^{(0)}, \bW^{(1)}, \bW^{\rm FC} \}$ as the parameters of the architecture transformer.
To cover all possible architectures, we randomly sample architectures from the whole architecture space and use them to train our model. 

\textbf{Differences with LSTM.}
The architecture graph can also be processed by the long short-term memory (LSTM)~\cite{hochreiter1997long}, which is a common practice in NAS methods~\cite{luo2018neural,zoph2016neural,pham2018efficient}. In these methods, LSTM first treats the graph as a sequence of tokens and then learns the information from the sequence. However, turning a graph into a sequence of tokens may lose some connectivity information of the graph, leading to limited performance. On the contrary, our GCN model can better exploit the information from the graph and yield superior performance (See results in Section~\ref{exp:derive_inference}). 
% and the length of the sequence would significantly influence the performance of LSTM (See results in Section~\ref{exp:derive_inference}). 

\subsection{Training and Inference of \sexyname}\label{sec:train}

\kuired{We apply the policy gradient~\cite{williams1992simple} to train our model. The overall scheme is shown in Algorithm~\ref{alg:training},  which employs an alternating manner.} Specifically, in each training epoch, \kuired{we first train the model parameters $w$ with fixed transformer parameters $\theta$. Then, we train the transformer parameters $\theta$ by fixing the model parameters $w$.} 

\begin{algorithm}[t]
\small
	\caption{Training method for Neural Architecture Transformer (\sexyname).}
	\label{alg:training}
	\begin{algorithmic}[1]
		\REQUIRE  The number of sampled input architectures in an iteration $m$, the number of sampled optimized architectures for each input architecture $n$, learning rate $\eta$, regularizer parameter $\lambda$ in Eqn.~(\ref{eq:obj_entropy}), input architecture distribution $p(\cdot)$, shared model parameters $w$, transformer parameters $\theta$.  \\
        \STATE Initiate $w$ and $\theta$. \\
		\WHILE{not convergent}
		\FOR{each iteration on training data} 
		\STATE // {Fix $\theta$ and update $w$.} \\
		\STATE Sample $ \beta_i \sim p(\cdot)$ to construct a batch $\{\beta_i \}_{i=1}^m$.\\
		\STATE Update the model parameters $w$ by descending the gradient:\\
		\STATE ~~~~~~~~$w \leftarrow w - \eta \frac{1}{m} \sum_{i=1}^{m} \nabla_{w} \mathcal{L}(\beta_i,w)$.\\
		\ENDFOR
		\FOR{each iteration on validation data}
		\STATE // {Fix $w$ and update $\theta$.} \\
		\STATE Sample $\beta_i \sim p(\cdot)$ to construct a batch $\{\beta_i \}_{i=1}^m$.\\
        \STATE Obtain $\{\alpha_j\}_{j=1}^n$ according to the policy learned by GCN.\\
		\STATE Update the transformer parameters $\theta$ by ascending the gradient:\\
% 		\STATE ~~~~~~~~$\theta \leftarrow \theta - \eta \frac{1}{M} \sum_{i=1}^{M} \Big( \nabla_{\theta} \log \pi(\alpha_i | \beta_i; \theta) \big( R( \alpha_i, w) - R(\beta_i, w) + \beta ~\nabla_{\theta} H \big(\pi(\beta_i; \theta) \big) \Big)$.\\
        \STATE ~~~~~~~~$\theta \leftarrow \theta + \eta \frac{1}{mn} \sum_{i=1}^{m} \sum_{j=1}^{n} \left[ \nabla_{\theta} \log \pi(\alpha_j | \beta_i; \theta) \big( R( \alpha_j, w) - R(\beta_i, w) \big) {+} \lambda \nabla_{\theta} H \big(\pi \left(\cdot|\beta_i; \theta \right) \big) \right] $.\\
		\ENDFOR
		\ENDWHILE
	\end{algorithmic}
\end{algorithm}

\textbf{Training the model parameters $w$.}
Given any $\theta$, we need to update the model parameters $w$ based on the training data. Here, to accelerate the training process, we adopt the parameter sharing technique~\cite{pham2018efficient}, \ie, 
we construct a large computational graph, where each subgraph represents a neural network architecture, hence forcing all architectures to share the parameters.
Thus, we can use the shared parameters $w$ to represent the parameters for different architectures. 
For any architecture $\beta \sim p(\cdot)$, let $\mathcal{L}(\beta,w)$ be the loss function on the training data, \eg, the cross-entropy loss. Then, given any $m$ sampled architectures, the updating rule for $w$ with parameter sharing can be given by $w \leftarrow w - \eta \frac{1}{m} \sum_{i=1}^{m} \nabla_{w} \mathcal{L}(\beta_i,w)$, where $\eta$ is the learning rate. 
% Once $w$ is updated, we will tend to update the transformer parameters $\theta$.

\textbf{Training the transformer parameters $\theta$.}
We train the transformer model with policy gradient~\cite{williams1992simple}. To encourage exploration, we introduce an entropy
regularization term into the objective to 
prevent the transformer from converging to a local optimum too quickly~\cite{zoph2018learning}, \eg, selecting the ``original'' option for all the operations.
Given the shared parameters $w$, the objective can be formulated as
% \begin{equation}\label{eq:obj_entropy}
% \begin{aligned}
% J(\theta) &= \mathbb{E}_{\alpha \sim \pi({\alpha|\beta};\theta), {\beta \sim p(\beta)}}~\left[ {R}\left(\alpha, w \right) - {R}\left(\beta, w \right)  + \beta H \big(\pi(\beta; \theta) \big) \right]\\
% &= \sum_{{\beta}} q({\beta}) \sum_\alpha \pi(\alpha | {\beta}; \theta) \big( R\left({\alpha}, w\right) - R({\beta}, w)  +  \beta H \big(\pi({\beta}; \theta) \big),
% \end{aligned}
% \end{equation}
\begin{equation}\label{eq:obj_entropy}
% \vspace{-20 pt}
\begin{aligned}
J(\theta) &= \mathbb{E}_{\beta \sim p(\cdot)} \left[ \mathbb{E}_{\alpha \sim \pi(\cdot |\beta;\theta)}  \left[ {R}\left(\alpha, w \right) - {R}\left(\beta, w \right) \right]  + \lambda H \big(\pi(\cdot|\beta; \theta) \big)  \right]\\
&= \sum_{\beta} p(\beta) \left[ \sum_\alpha \pi(\alpha | \beta; \theta) \big( {R}\left(\alpha, w \right) - {R}\left(\beta, w \right)\big) +  \lambda H \big(\pi(\cdot|{\beta}; \theta) \big) \right].
\end{aligned}
% \vspace{-10 pt}
\end{equation}
where $p(\beta)$ is the probability to sample some architecture $\beta$ from the distribution $p(\cdot)$, $\pi(\alpha|\beta;\theta)$ is the probability to sample some architecture $\alpha$ from the distribution $\pi({\cdot|\beta};\theta)$, $H(\cdot)$ evaluates the entropy of the policy, and $\lambda$ controls the strength of the entropy regularization term.
% Usually, a baseline $R(\beta, w)$ is subtracted from the reward to reduce the variance of gradient estimate~\cite{li2017deep}. 
% Based on Eqn.~(\ref{eq:obj_entropy}), 
For each input architecture, we sample $n$ optimized architectures $\{\alpha_j\}_{j=1}^{n}$ from the distribution $\pi(\cdot | \beta; \theta)$ in each iteration.
Thus, the gradient of Eqn.~(\ref{eq:obj_entropy}) w.r.t. $\theta$ becomes\footnote{We put the derivations of Eqn.~(\ref{eq:entropy_gradient}) in the supplementary.}
% \begin{equation}\label{eq:entropy_gradient}
%     \nabla_{\theta} J(\theta) \approx \frac{1}{M} \sum_{i=1}^{M} \nabla_{\theta} \log \pi(\alpha_i | \beta_i; \theta) \big( R(\alpha_i, w) - R(\beta_i, w) \big) + \beta \nabla_{\theta} H \big(\pi(\beta_i; \theta) \big).
% \end{equation}
\begin{equation}\label{eq:entropy_gradient}
    \nabla_{\theta} J(\theta) \approx \frac{1}{mn} \sum_{i=1}^{m} \sum_{j=1}^{n} \left[ \nabla_{\theta} \log  \pi(\alpha_j | \beta_i; \theta) \big( R(\alpha_j, w) - R(\beta_i, w) \big) + \lambda \nabla_{\theta} H \big(\pi(\cdot|\beta_i; \theta) \big) \right].
\end{equation}
The regularization term $H \big(\pi(\cdot|\beta_i; \theta) \big)$ encourages the distribution $\pi(\cdot|\beta; \theta)$ to have high entropy, 
% The regularization term encourages $\alpha$ to have high entropy, 
\ie, high diversity in the decisions on the edges. Thus, the decisions for some operations would be encouraged to choose the ``identity'' or ``null'' operations during training. As a result, \sexyname is able to sufficiently explore the whole search space to find the optimal architecture.
% \guo{It is worth noting that the whole training process of \sexyname only takes 0.17 GPU day based on a Tesla P40 GPU.}

\textbf{Inferring the optimized architecture.}
\kuired{We do not explicitly obtain the optimized architecture via $\alpha = {\rm NAT} (\beta; \theta)$.}  Instead, we conduct sampling according to the learned probability distribution.
Specifically, we first sample several candidate optimized architectures from the learned policy $\pi(\cdot|\beta; \theta)$ and then select the architecture with the highest validation accuracy.
% \footnote{Note that the performance of different architectures based on the shared parameters is in good consistency with the final performance of these architectures~\cite{pham2018efficient}.}.
Note that we can also obtain the optimized architecture by selecting the operation with the maximum probability, which, however, tends to reach a local optimum and yields worse results than the sampling based method (See comparisons in Section~\ref{exp:derive_inference}).

\section{Experiments}\label{sec:exp}

In this section, we apply \sexyname on both hand-crafted and NAS based architectures, and conduct experiments on two \kuired{image classification benchmark datasets,} \ie, CIFAR-10~\citep{krizhevsky2009learning} and ImageNet~\citep{deng2009imagenet}.
% and compare the performance on two benchmark image classification datasets, \ie, CIFAR-10~\citep{krizhevsky2009learning} and ImageNet~\citep{deng2009imagenet}.
All implementations are based on PyTorch.\footnote{\kuired{The source code of \sexyname} is available at \href{https://github.com/guoyongcs/NAT}{https://github.com/guoyongcs/NAT}.}

\subsection{Implementation Details}\label{sec:implementation}
We consider two kinds of cells in a deep network, including the normal cell and the reduction cell.
The normal cell preserves the same spatial size as inputs while the reduction cell reduces the spatial size by $2\times$. Both the normal and reduction cells contain 2 input nodes and a number of intermediate nodes.
During training, we build the deep network by stacking 8 basic cells and train the transformer for 100 epochs. 
% We set the number of sampled architectures to $M=1$ 
We set $m=1$, $n=1$, and $\lambda=0.003$ in the training.
We split CIFAR-10 training set into $40\%$ and $60\%$ slices to train the model parameters $w$ and the transformer parameters $\theta$, respectively.
As for the evaluation of the networks with different architectures, 
we replace the original cell with the optimized one and train the model from scratch. Please see more details in the supplementary.
\guo{For all the considered architectures, we follow the same settings of the original papers. In the experiments, we only apply cutout to the NAS based architectures on CIFAR-10.}
	
\begin{table}[t]
    \centering
    \caption{Performance comparisons of the optimized architectures obtained by different methods based on hand-crafted architectures. ``/'' denotes the original models that are not changed by architecture optimization methods.
    }
    \resizebox{1.0\textwidth}{!}{
        \begin{tabular}{ccccc|ccccccc}
        \toprule[1pt]
        \multicolumn{5}{c|}{\multirow{1}[0]{*}{CIFAR-10}} & \multicolumn{6}{c}{\multirow{1}[0]{*}{ImageNet}}\\
        \hline
        \multicolumn{1}{c}{\multirow{2}[0]{*}{Model}}  & \multicolumn{1}{c}{\multirow{2}[0]{*}{Method}} & \multicolumn{1}{c}{\multirow{2}[0]{*}{\#Params (M)}} & \multicolumn{1}{c}{\multirow{2}[0]{*}{\#MAdds (M)}} & 
        \multicolumn{1}{c|}{\multirow{2}[0]{*}{Acc. (\%)}} &
        \multicolumn{1}{c}{\multirow{2}[0]{*}{Model}}  & \multicolumn{1}{c}{\multirow{2}[0]{*}{Method}} & \multicolumn{1}{c}{\multirow{2}[0]{*}{\#Params (M)}} & \multicolumn{1}{c}{\multirow{2}[0]{*}{\#MAdds (M)}} & 
        \multicolumn{2}{c}{Acc. (\%)} \\
        \cline{10-11}
        & & & & & & & & &  \multicolumn{1}{l}{Top-1} & \multicolumn{1}{l}{Top-5} \\
        \hline
        \multirow{3}[0]{*}{VGG16}  & /    &   15.2    &   313   & 93.56 &  \multirow{3}[0]{*}{VGG16}  & / &   \multirow{1}[0]{*}{138.4}    &  \multirow{1}[0]{*}{15620}     & 71.6 & 90.4 \\
        & NAO\cite{luo2018neural}   &   19.5    &   548   & 95.72 &   &  NAO~\cite{luo2018neural}  & 147.7 & 18896     & 72.9 & 91.3\\
        & \sexyname    &   15.2    &   315   & \textbf{96.04} & 
         &  \sexyname &   138.4   &    15693   & \textbf{74.3} & \textbf{92.0}\\
        \hline
        \multirow{3}[0]{*}{ResNet20} & /    &   0.3    &   41   & 91.37 & \multirow{3}[0]{*}{ResNet18} & /  &   \multirow{1}[0]{*}{11.7}     &  \multirow{1}[0]{*}{1580}     & 69.8 & 89.1 \\
        & NAO~\cite{luo2018neural}      &   0.4    &   61   & 92.44  &  &  NAO~\cite{luo2018neural}   &   17.9     &  2246     & 70.8 & 89.7\\
        & \sexyname     &   0.3    &   42   & \textbf{92.95} &  &  \sexyname  &    11.7    &  1588 &  \textbf{71.1}  & \textbf{90.0}  \\
        \hline
        \multirow{3}[0]{*}{ResNet56} & /      &   0.9    &   127   & 93.21 & \multirow{3}[0]{*}{ResNet50} &   /  &  \multirow{1}[0]{*}{25.6}     &  \multirow{1}[0]{*}{3530}     & 76.2 & 92.9\\
        & NAO~\cite{luo2018neural}       &   1.3    &   199   & 95.27  & &  NAO~\cite{luo2018neural}     &  34.8     &  4505     & 77.4 & 93.2 \\
        & \sexyname     &   0.9    &   129   & \textbf{95.40} &  &  \sexyname &  25.6    &   3547    & \textbf{77.7} & \textbf{93.5} \\
        \hline
        \multirow{3}[0]{*}{MobileNetV2} & /      &   2.3    &  91   & 94.47 & \multirow{3}[0]{*}{MobileNetV2} &   /  &  \multirow{1}[0]{*}{3.4}     &  \multirow{1}[0]{*}{300}     & 72.0 & 90.3 \\
        & NAO~\cite{luo2018neural}       &   2.9    &   131   & 94.75  & &  NAO~\cite{luo2018neural}     &  4.5     &  513     & 72.2 & 90.6 \\
        & \sexyname     &   2.3    &   92   & \textbf{95.17} &  &  \sexyname &  3.4    &   302    & \textbf{72.5} & \textbf{91.0} \\
        \bottomrule[1pt]
        \end{tabular}
        }
        \label{tab:hand-crafted}
\end{table}

\begin{figure*}[t]
	\centering
	\includegraphics[width=0.9\columnwidth]{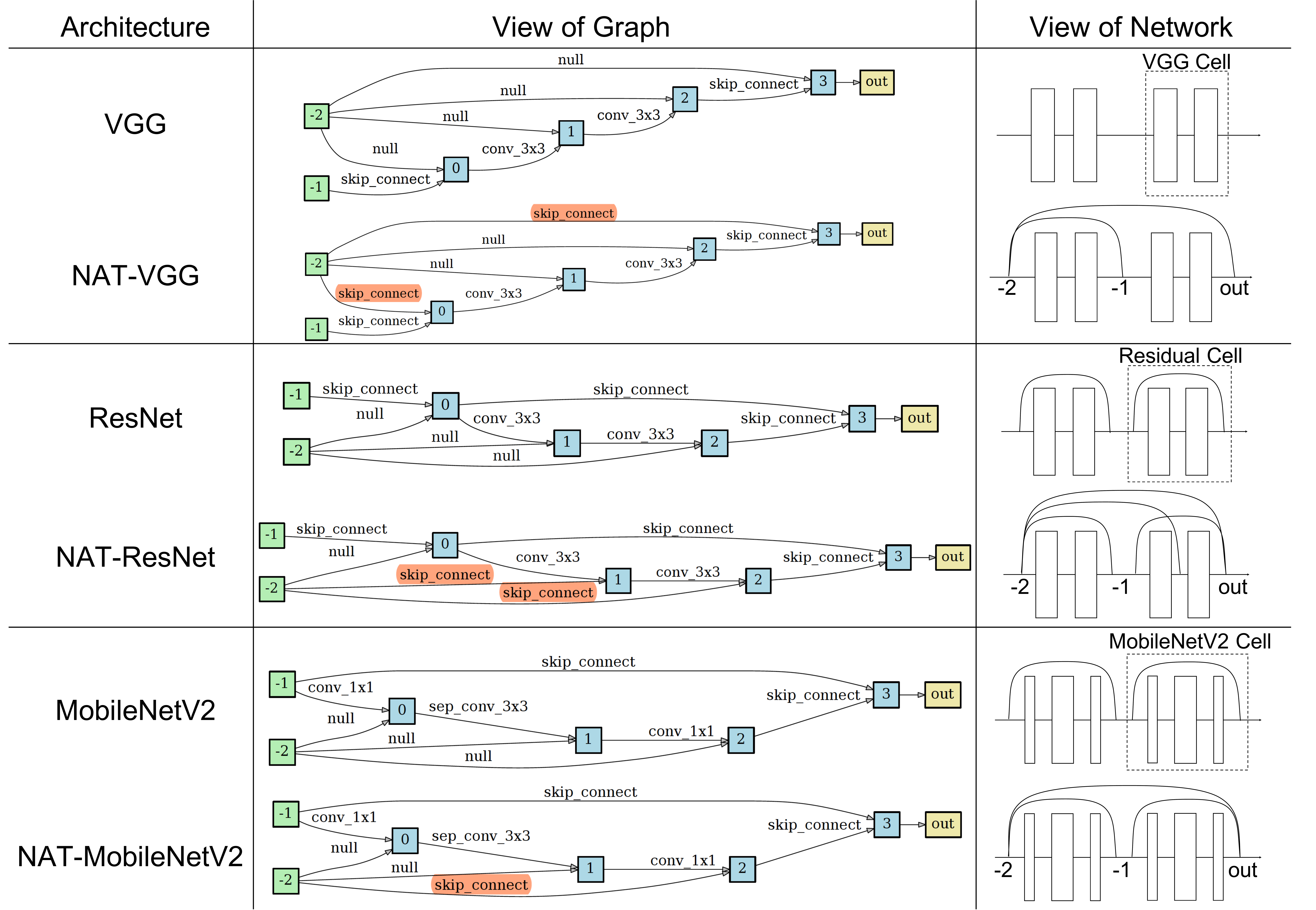}
	\caption{Architecture optimization results of hand-crafted architectures. We provide both the views of graph (left) and network (right) to show the differences in architecture.
	}
	\vspace{-10 pt}
	\label{fig:vgg_resnet}
\end{figure*}

\begin{table}[t!]
    \centering
    \caption{Comparisons of the optimized architectures obtained by different methods based on NAS based architectures. ``-'' denotes that the results are not reported. ``/'' denotes the original models that are not changed by architecture optimization methods. $^\dagger$ denotes the models trained with cutout.
    }
    \resizebox{1.0\textwidth}{!}{
        \begin{tabular}{ccccc|ccccccc}
        \toprule[1pt]
        \multicolumn{5}{c|}{\multirow{1}[0]{*}{CIFAR-10}} & \multicolumn{6}{c}{\multirow{1}[0]{*}{ImageNet}}\\
        \hline
        \multicolumn{1}{c}{\multirow{2}[0]{*}{Model}}  & \multicolumn{1}{c}{\multirow{2}[0]{*}{Method}} & \multicolumn{1}{c}{\multirow{2}[0]{*}{\#Params (M)}} & \multicolumn{1}{c}{\multirow{2}[0]{*}{\#MAdds (M)}} & 
        \multicolumn{1}{c|}{\multirow{2}[0]{*}{Acc. (\%)}} &
        \multicolumn{1}{c}{\multirow{2}[0]{*}{Model}}  & \multicolumn{1}{c}{\multirow{2}[0]{*}{Method}} & \multicolumn{1}{c}{\multirow{2}[0]{*}{\#Params (M)}} & \multicolumn{1}{c}{\multirow{2}[0]{*}{\#MAdds (M)}} & 
        \multicolumn{2}{c}{Acc. (\%)} \\
        \cline{10-11}
        & & & & & & & & &  \multicolumn{1}{l}{Top-1} & \multicolumn{1}{l}{Top-5} \\
        \hline
        AmoebaNet$^\dagger$~\cite{real2018regularized} & \multirow{4}[0]{*}{/} & 3.2 & - &  96.73 & AmoebaNet~\cite{real2018regularized} & \multirow{4}[0]{*}{/} & 5.1 & 555 &  74.5 & 92.0 \\
        PNAS$^\dagger$~\cite{liu2018progressive} &  & 3.2 & - & 96.67 & PNAS~\cite{liu2018progressive} &  & 5.1 & 588 &  74.2 & 91.9 \\
        SNAS$^\dagger$~\cite{xie2018snas} &  & 2.9 & - & 97.08 & SNAS~\cite{xie2018snas} &  & 4.3 & 522 &  72.7 & 90.8 \\
        GHN$^\dagger$~\cite{zhang2018graph} &  & 5.7 & - & 97.22 & GHN~\cite{zhang2018graph} &  & 6.1 & 569 &  73.0 & 91.3 \\
        \hline
        \multirow{3}[0]{*}{ENAS$^\dagger$~\cite{pham2018efficient}}  & /     &   4.6    &   804   & 97.11 & \multirow{3}[0]{*}{ENAS~\cite{pham2018efficient}} & /  &    5.6   &   607    & 73.8  & 91.7 \\
        & NAO~\cite{luo2018neural}      &   4.5    &   763   & 97.05  &  &  NAO~\cite{luo2018neural}    &    5.5   &   589   & 73.7 & 91.7 \\
        & \sexyname     &  4.6    &    804   & \textbf{97.24} &  &  \sexyname & 5.6   &   607    & \textbf{73.9} & \textbf{91.8} \\
        \hline
        \multirow{3}[0]{*}{DARTS$^\dagger$~\cite{liu2018darts}}  & /   &   3.3    &   528   & 97.06 & \multirow{3}[0]{*}{DARTS~\cite{liu2018darts}} & /  &    4.7   &   574    & 73.1 & 91.0\\
        & NAO~\cite{luo2018neural}   &   3.5    &   577   & 97.09 &  &  NAO~\cite{luo2018neural}     &    5.1   &   627    & 73.3 & 91.1  \\
        & \sexyname    &   {2.7}    &   424    &  \textbf{97.28} & &  \sexyname  & 4.0    &   441    & \textbf{73.7} & \textbf{91.4} \\
        \hline
        \multirow{3}[0]{*}{NAONet$^\dagger$~\cite{luo2018neural}}  & /   &   128    &   66016   & 97.89 & \multirow{3}[0]{*}{NAONet~\cite{luo2018neural}} & /  &    11.35   &   1360    & 74.3 & 91.8 \\
        & NAO~\cite{luo2018neural}   &   143    &  73705    & 97.91 &  &  NAO~\cite{luo2018neural}     &  11.83     &  1417     & 74.5  & 92.0  \\
        & \sexyname    &         113    & 58326  & \textbf{98.01} & &  \sexyname  &  8.36     &  1025     & \textbf{74.8} & \textbf{92.3} \\
        \bottomrule[1pt]
        \end{tabular}
        }
        \label{tab:NAS-based}%
\end{table}

\begin{figure*}[t!]
	\centering
	\vspace{-10 pt}
    \includegraphics[width=0.9\columnwidth]{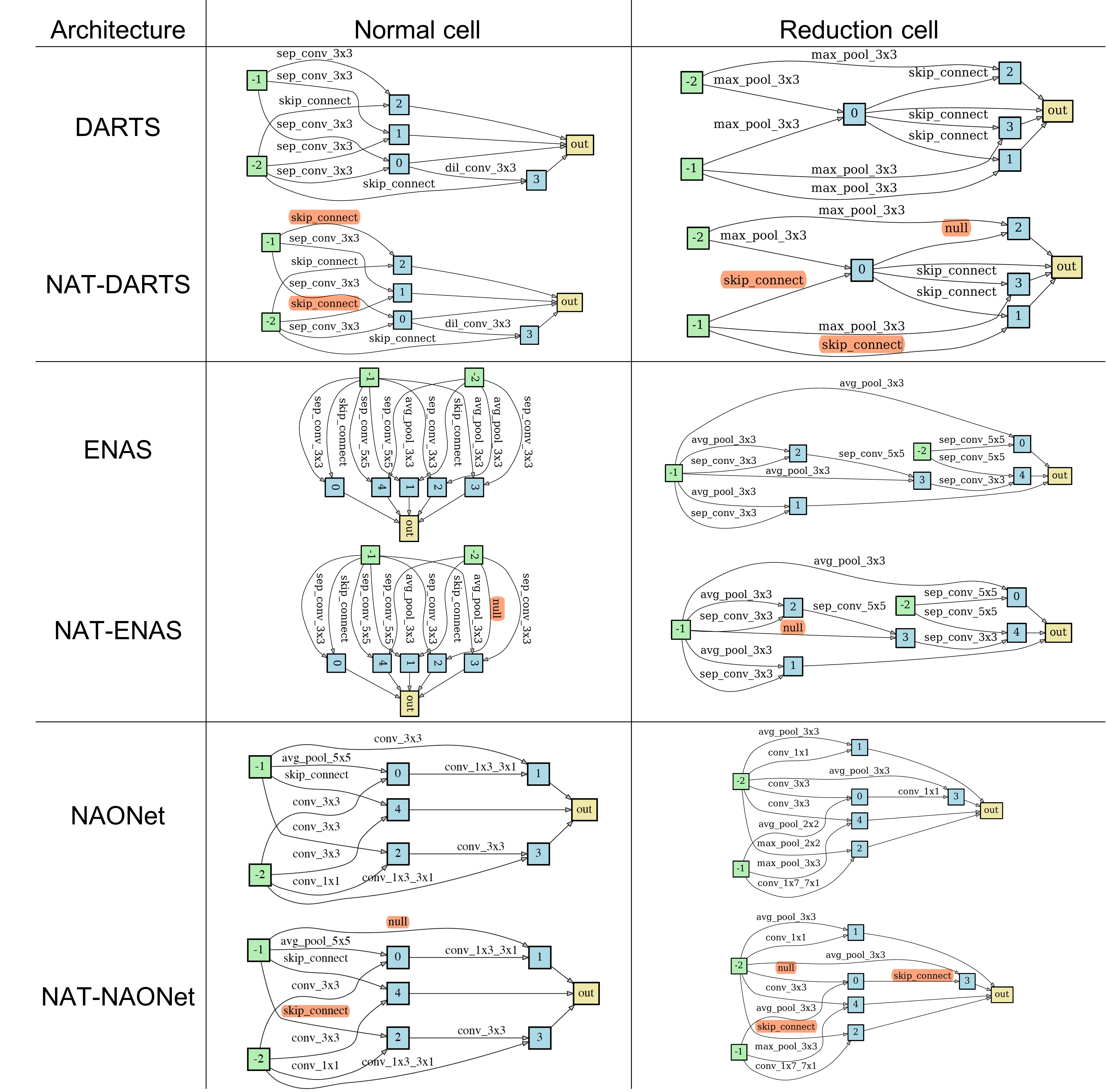}
	\caption{Architecture optimization results on the architectures of NAS based architectures.
	}
	\vspace{-15 pt}
	\label{fig:darts_enas}
\end{figure*}

\subsection{Results on Hand-crafted Architectures}
    
    In this experiment, we apply \sexyname on three popular hand-crafted models, \ie, VGG~\cite{simonyan2014very}, ResNet~\cite{he2016deep}, and MobileNet~\cite{sandler2018mobilenetv2}.
    % which correspond to the plain and residual architectures, respectively.
    \guo{To make all architectures share the same graph representation method defined in Section~\ref{sec:MDP}, we add null connections into the hand-crafted architectures to ensure that each node has two input nodes (See examples in Figure~\ref{fig:vgg_resnet}).}
    For a fair comparison, we build deep networks using the original and optimized architectures while keeping the same depth and number of channels as the original models.
    We compare \sexyname with a strong baseline method Neural Architecture Optimization (NAO)~\cite{luo2018neural}.
    We show the results in Table~\ref{tab:hand-crafted} and the corresponding architectures in Figure~\ref{fig:vgg_resnet}. 
     From Table~\ref{tab:hand-crafted}, although the models with NAO yield better performance than the original ones, 
    they often have more parameters and higher computation cost. By contrast, 
    our \sexyname based models consistently outperform the original models by a large margin with approximately the same computation cost.

\subsection{Results on NAS Based Architectures}\label{exp:nas}

For the automatically searched architectures, 
we evaluate the proposed \sexyname on three state-of-the-art NAS based architectures, \ie, DARTS~\cite{liu2018darts}, NAONet~\cite{luo2018neural}, and ENAS~\cite{pham2018efficient}. 
Moreover, we also compare our optimized architectures with other NAS based architectures, including AmoebaNet~\cite{real2018regularized}, PNAS~\cite{liu2018progressive}, SNAS~\cite{xie2018snas}
and GHN~\cite{zhang2018graph}. 
From Table~\ref{tab:NAS-based}, 
% both \sexyname-ENAS and \sexyname-DARTS 
all the NAT based architectures
yield higher accuracy than their baseline models and the models optimized by NAO on CIFAR-10 and ImageNet. 
Compared with other NAS based architectures, our \sexyname-DARTS performs the best on CIFAR-10 and achieves the competitive performance compared to the best architecture (\ie, AmoebaNet) on ImageNet with less computation cost and fewer number of parameters.
We also visualize the architectures of the original and optimized cell 
in Figure~\ref{fig:darts_enas}. 
As for DARTS and NAONet, \sexyname replaces several redundant operations with the skip connections or directly removes the connection, leading to fewer number of parameters. While optimizing ENAS, \sexyname removes the average pooling operation and
improves the performance without introducing extra computations.
% Due to the page limit, we show the architectures optimized by NAO in the supplementary.

\subsection{Comparisons of Different Policy Learners}\label{exp:derive_inference}

% Based on CIFAR-10, we study the performance with different policy learners, namely as Random Search, LSTM, and GCN on . 

% , and different inference methods, \ie, sampling based and maximum-probability based methods.

\guocam{In this experiment, we compare the performance of different policy learners, including Random Search, LSTM, and the GCN method.}
% \qicam{We conduct more experiments on CIFAR-10 to compare the performance with different policy learners, including Random Search, LSTM, sampling based and maximum probability based methods.}
%  to compare the perform
For the Random Search method, we perform random transitions among $O$, $S$, and $N$ on the input architectures.
\guocam{For the GCN method, we consider two variants which infer the optimized architecture by sampling from the learned policy (denoted by Sampling-GCN) or by selecting the operation with the maximum probability (denoted by Maximum-GCN).
% We show the results in Table \ref{tab:sup-handcrafted}.
}
% From Table \ref{tab:sup-handcrafted},
% Maximum-GCN represents to derive the architecture by selecting the operation with the maximum probability,
% and Sampling-GCN represents conducting sampling according to the probability distribution.
From Table \ref{tab:sup-handcrafted},
our Sampling-GCN method outperforms all the considered policies on different architectures. These results demonstrate the superiority of the proposed GCN method as the policy learner.

\subsection{Effect of Different Graph Representations \kuired{on} Hand-crafted Architectures}
In this experiment, we investigate the effect of different graph representations on hand-crafted architectures.
Note that an architecture may correspond to many different topological graphs, especially for the hand-crafted architectures, \eg, VGG and ResNet, where the number of nodes is smaller than that of our basic cell. 
For convenience, we study three different graphs for VGG and ResNet20, respectively. 
The average accuracy of \sexyname-VGG is 95.83\% and outperforms the baseline VGG with the accuracy of 93.56\%. 
Similarly, our \sexyname-ResNet20 yields the average accuracy of 92.48\%, which is also better than the original model.
We put the architecture and the performance of each possible representation in the supplementary.
In practice, the graph representation may influence the result of \sexyname and how to alleviate its effect still remains an open question.

\begin{table}[t]
  \centering
  \caption{Performance comparisons of the architectures obtained by different methods on CIFAR-10. The reported accuracy (\%) is the average performance of five runs with different random seeds. ``/'' denotes the original models that are not changed by architecture optimization methods. $^\dagger$ denotes the models trained with cutout.}
  \small
  \resizebox{0.85\textwidth}{!}
  {
    \begin{tabular}{c|ccccccc}
    \toprule[1pt]
    Method & \multicolumn{1}{c}{VGG16} & \multicolumn{1}{c}{ResNet20} & \multicolumn{1}{c}{MobileNetV2} & \multicolumn{1}{c}{ENAS$^\dagger$} & \multicolumn{1}{c}{DARTS$^\dagger$}  & \multicolumn{1}{c}{NAONet$^\dagger$}  \\
    \hline 
    /  &    93.56   &   91.37    &  94.47       & 97.11 & 97.06 & 97.89 \\
    Random Search & 93.17 & 91.56 & 94.38 & 96.58 & 95.17  & 96.31  \\
    LSTM &   94.45   &   92.19    &  95.01 &   97.05 & 97.05  & 97.93 \\
    Maximum-GCN &    94.37  &   92.57  & 94.87 & 96.92 & 97.00  & 97.90  \\
    Sampling-GCN (Ours) &   \textbf{95.93}    &   \textbf{92.97}    &    \textbf{95.13}      & \textbf{97.21} & \textbf{97.26} & \textbf{97.99} \\
    \bottomrule[1pt]
    \end{tabular}%
    }
  \label{tab:sup-handcrafted}%
  \vspace{-10 pt}
\end{table}%

\section{Conclusion}

In this paper, we \kuired{have proposed} a novel Neural Architecture \kuired{Transformer} (\sexyname) \kuired{for the task of architecture optimization}. To solve this problem, we cast it into a Markov decision process (MDP) by making a series of decisions to optimize existing operations with more computationally efficient operations, including skip connection and null operation. To show the effectiveness of \sexyname, we apply it to both hand-crafted architectures and Neural Architecture Search (NAS) based architectures. Extensive experiments on CIFAR-10 and ImageNet datasets demonstrate the effectiveness of the proposed method in improving the accuracy and the compactness of neural architectures.

\subsubsection*{Acknowledgments}
This work was partially supported by Guangdong Provincial Scientific and Technological Funds under Grants 2018B010107001, National Natural Science Foundation of China (NSFC) (No. 61602185), key project of NSFC (No. 61836003), Fundamental Research Funds for the Central Universities (No. D2191240), Program for Guangdong Introducing Innovative and Enterpreneurial Teams 2017ZT07X183,  Tencent AI Lab Rhino-Bird Focused Research Program (No. JR201902), Guangdong Special Branch Plans Young Talent with Scientific and Technological Innovation (No. 2016TQ03X445), Guangzhou Science and Technology Planning Project (No. 201904010197), and Microsoft Research Asia (MSRA Collaborative Research Program). We last thank Tencent AI Lab.

% MSRA Collaborative Research (No. x2rjD5190340).
% This work was partially supported by National Natural Science Foundation of China (NSFC) (No. 61602185, 61876208, and 61502177), Key Project of NSFC (No. 61836003),
% Program for Guangdong Introducing Innovative and Enterpreneurial Teams (No. 2017ZT07X183), Guangdong Provincial Scientific and Technological Funds (No. 2018B010107001),
% Tencent AI Lab Rhino-Bird Focused Research Program (No. JR201902), Guangdong Special Branch Plans Young Talent with Scientific and Technological Innovation (No. 2016TQ03X445), 
% Guangzhou Science and Technology Planning Project (No. 201904010197). 
% We thank the Tencent Seven Team for providing GPU resources.
%We also thank the anonymous reviewers.

\bibliographystyle{abbrv}
{
	\small
	\bibliography{nips_ref}
}

\end{document}